\pdfoutput=1 
\documentclass[10pt, a4paper]{article}

\usepackage[hyphens]{url}
\usepackage[final]{lrec2026}
\usepackage{subfiles}

\title{UIC-AIHealth4All at ArchEHR-QA 2026: Answer-First Evidence Grounding for Clinical Question Answering\thanks{Accepted at CL4Health @ LREC 2026. This is the authors' version; the version of record will appear in the ACL Anthology.}}

\name{Mohammad Arvan, Hossein Haeri, Natalie Parde, Rebecca T. Feinstein}

\address{University of Illinois Chicago \\
         Chicago, IL, USA \\
         \{marvan3, haeri, parde, rfeinst\}@uic.edu}

\abstract{
We describe the UIC-AIHealth4All system for ArchEHR-QA 2026, a shared task on grounded question answering from electronic health records. We participated in Subtasks~2 (evidence identification), 3 (answer generation), and 4 (answer-evidence alignment). For Subtasks~2 and 3, we propose an \textit{answer-first} pipeline in which the model generates candidate answers citing specific note sentences before classifying the full evidence set, exploiting the asymmetry between judging relevance in the abstract versus relative to a generated answer. For Subtask~4, we apply self-consistency voting over five independent model calls, retaining links by vote threshold. Our pipeline ranked third on evidence identification (Strict Micro F1 62.90), ninth on answer generation (Overall 31.90), and fifth on answer-evidence alignment (F1 79.81). A post-hoc linguistic analysis of 45 stylistic features reveals that model outputs remain 3.2 Flesch-Kincaid grade levels harder to read than clinician-authored references despite matching their word and sentence counts, suggesting readability warrants explicit optimization in clinical NLP systems. Code and prompts are available at \url{https://github.com/mo-arvan/archehr-qa-2026-uic-aihealth4all}.
\\ \newline \Keywords{clinical NLP, question answering, electronic health records, evidence identification, grounded generation}}

\begin{document}

\maketitleabstract

\section{Introduction}

Grounding an answer means anchoring every claim to a specific passage in the source document. In safety-critical domains such as clinical question answering, grounding is a reliability requirement: an answer that cannot be verified against the source text cannot be trusted for medical decisions. This requirement creates a joint constraint on evidence selection and answer generation.  The minimal evidence set depends on what the answer will say, and a faithful answer cannot exceed what the evidence supports.

Grounded QA from electronic health records (EHRs), where both operations must be solved jointly, has received less attention than individually scoped clinical QA tasks~\cite{soni-etal-2025-archehr-qa}. Prior work has addressed factoid retrieval from structured data~\cite{DBLP:conf/nips/LeeHBKSYSKC22}, reading comprehension over published literature~\cite{jin-etal-2019-pubmedqa}, and information extraction from clinical text~\cite{uzuner20112010}, each formulated as an independent subtask. Whether generating a candidate answer before classifying evidence improves the accuracy of evidence identification has not been established.

ArchEHR-QA 2026~\cite{soni-etal-2026-archehr-qa,soni-demner-fushman-2026-dataset} directly addresses grounded EHR question answering across four subtasks: question interpretation (Subtask~1), evidence identification (Subtask~2), answer generation (Subtask~3), and answer-evidence alignment (Subtask~4). The dataset is derived from clinical notes in the MIMIC-III~\cite{PhysioNet-mimiciii-1.4} and MIMIC-IV~\cite{PhysioNet-mimiciv-2.2} databases, with patient-authored questions, clinician-interpreted reformulations, sentence-level relevance annotations, and clinician-authored reference answers with sentence-level evidence alignments. The shared task provides a setting in which all three components of the grounded QA pipeline (evidence, answer, and alignment) can be evaluated against expert clinical judgments.

We participated in three of the four subtasks (Subtasks~2, 3, and 4) as the UIC-AIHealth4All team. Our central contribution is an answer-first pipeline for evidence identification and answer generation (Subtasks~2 and 3), extending our prior system~\cite{arvan-etal-2025-uic}, in which the model first generates one or more candidate answers citing specific note sentences, then classifies the full evidence set in light of those candidates, and finally produces a submission answer from the classified evidence. For Subtask~4, we applied self-consistency voting over multiple independent model calls, aggregating link predictions by vote count and selecting a threshold that balanced precision and recall on the development set.

We additionally characterized stylistic differences between model outputs and clinician-authored reference answers. We computed 45 linguistic features across six categories (lexical, syntactic, readability, stylistic, clinical, and sentiment) and compared them with paired Wilcoxon signed-rank tests. This linguistic analysis guided prompt refinement and revealed a residual register gap between model outputs and clinician answers that persists after length and sentence count are matched.

Our main results are as follows: (1) the answer-first pipeline ranked third on evidence identification (Strict Micro F1 62.90); (2) our answer generation submission ranked ninth overall (test score 31.90), with iterative constraint enforcement achieving 100\% output compliance on both development and test sets within five correction attempts; (3) self-consistency voting ranked fifth on answer-evidence alignment (F1 79.81); and (4) despite matching clinician word count and sentence count, model outputs remained 3.2 Flesch-Kincaid grade levels harder to read than the clinician reference. These results suggest that answer-first ordering offers a practical advantage for evidence identification and that readability warrants explicit optimization in clinical NLP systems serving lay users.

\section{Method}

\subsection{Study Design and Setting}

We participated in Subtasks~2, 3, and 4 of ArchEHR-QA 2026~\cite{soni-etal-2026-archehr-qa}, a shared task on grounded question answering from clinical notes. Subtask~2 (evidence identification) asks a system to select the minimal set of note sentences that support answering the patient question. Subtask~3 (answer generation) asks a system to generate a natural-language answer of at most 75 words, grounded in those sentences. Subtask~4 (answer-evidence alignment) asks a system to map each sentence of a pre-written reference answer to the specific note sentences that support it.

\subsection{Data}

All experiments used the ArchEHR-QA 2026 development and test sets, derived from the MIMIC-III~\cite{PhysioNet-mimiciii-1.4} and MIMIC-IV~\cite{PhysioNet-mimiciv-2.2} clinical databases. Each case contains a patient-authored question, a clinician-interpreted reformulation (at most 15 words), a clinical note excerpt with sentence-level relevance annotations, and a clinician-authored reference answer with sentence-level evidence alignments. The development set contains 20 cases. The test set for Subtasks~2 and 3 contains 47 cases; the test set for Subtask~4 contains 147 cases.

\subsection{Systems}

We designed two systems building on our prior work~\cite{arvan-etal-2025-uic}: an answer-first pipeline for Subtasks~2 and 3, and a self-consistency voting pipeline for Subtask~4. Code and prompts are available at \url{https://github.com/mo-arvan/archehr-qa-2026-uic-aihealth4all}.

All experiments used GPT-5.1 and GPT-5.2 through the Azure OpenAI API with structured outputs (\texttt{responses.parse}) and Pydantic schema validation. Both models support a reasoning effort parameter; we swept over \textit{medium} and \textit{high} settings during development and used \textit{medium} for all test submissions. The Subtask~2 ensemble spanned both model versions and multiple $k$ settings; Subtask~4 used GPT-5.2. No local GPU compute was required.

\textbf{Answer-first pipeline (Subtasks~2 and 3).} We designed a modular pipeline that jointly produced evidence predictions for Subtask~2 and an answer for Subtask~3 in three sequential stages.

\textbf{Stage 1: candidate answer generation.} We prompted the model to generate $k$ independent candidate answers, each citing specific note sentences by ID. The full note, the patient question, and the clinician-interpreted question were provided as input. The generation prompt specified past tense for events during the hospitalization and present tense for current state and prognosis, required third-person subject constructions (``The patient''), and targeted a word count of 70 to 75 words.

\textbf{Stage 2: answer-grounded evidence classification.} We presented the $k$ candidates alongside the clinical note and instructed the model to classify each note sentence as \textit{Essential}, \textit{Supplementary}, or \textit{Not Relevant}, using a minimal-set criterion: a sentence was Essential only if removing it would leave the generated answers unsupported. We refer to this method as the \textit{answer-first} pipeline. It exploits an asymmetry in difficulty: judging which sentences the model relied upon after generating an answer is easier than judging relevance in the abstract.

\textbf{Stage 3: final answer generation.} We instructed the model to generate the submission answer using only Essential (and optionally Supplementary) sentences, subject to the 75-word constraint. Each response was validated against the word limit, citation consistency, and output format. Constraint violations triggered a correction call that named the specific error and included the previous response as context, with up to five attempts per case. We logged all constraint violations by error type across all experiments to characterize the dominant failure modes of the pipeline.

For our best Subtask~2 submission, we combined seven configurations by majority vote: a note sentence was predicted Essential if at least four of the seven runs classified it as such. Ensemble members spanned both model versions and multiple $k$ settings to maximize prediction diversity.

\textbf{Self-consistency voting pipeline (Subtask~4).} We designed a self-consistency voting pipeline for Subtask~4 in which $N = 5$ independent model calls per case were aggregated by vote count to produce the final set of answer-sentence to note-sentence link predictions.

\textbf{Prompt design.} The alignment prompt instructed the model to apply a precision-first principle: cite only the note sentence that directly states the specific fact in the answer sentence. The prompt documented four common alignment errors with corrective examples: (1) over-citing, where contextual sentences are cited instead of the direct-claim sentence; (2) under-citing, where distinct claims sharing a source sentence are omitted; (3) citing synthesized conclusions, where no single note sentence directly states the inference; and (4) citing structural text such as section headers or symptom checklists. A worked example from the task documentation illustrated correct and incorrect alignments for each error type.

\textbf{Self-consistency voting.} Each of the $N = 5$ samples produced a set of (answer sentence, note sentence) link pairs. A link was included in the final prediction if it appeared in at least $t$ of the 5 samples. We selected $t = 3$ based on a precision-recall trade-off analysis on the development set.

\subsection{Measures and Outcomes}

We evaluated system outputs using the official task metrics and supplementary compliance measures.

For Subtask~2, gold labels classify each sentence as \textit{essential}, \textit{supplementary}, or \textit{not-relevant}. We report precision, recall, and F1 against \textit{essential} sentences (strict) and against the union of essential and supplementary (lenient). We additionally measured first-attempt constraint compliance as the percentage of outputs satisfying all format and length requirements on the first generation attempt, and logged all violations by error type.

For Subtask~3, we report lexical overlap (BLEU, ROUGE), semantic similarity (BERTScore, AlignScore), medical concept coverage (MEDCON), and simplification quality (SARI).

For Subtask~4, we report precision, recall, and F1 over predicted alignment links.

\subsection{Statistical and Analytic Methods}

To characterize how closely model outputs matched the stylistic register of clinician-authored reference answers, we computed 45 linguistic features for each of the 20 development cases and compared them against the clinician references. Features spanned six categories: lexical (word count, type-token ratio, Measure of Textual Lexical Diversity (MTLD), hapax ratio, average word length), syntactic (sentence count, average sentence length, parse depth, part-of-speech (POS) ratios), readability (Flesch-Kincaid Grade, Coleman-Liau Index, Flesch Reading Ease), stylistic (function-word percentage, passive-voice ratio, pronoun usage, frequency of the phrase \textit{the patient}), sentiment and tone (polarity, subjectivity, hedging and warning counts), and clinical communication (certainty markers, temporal and causal connectives, medical abbreviation count).

We applied paired Wilcoxon signed-rank tests to identify features that systematically diverged between model and clinician outputs. This non-parametric test was chosen because the 20-case sample cannot be assumed to yield normally distributed feature differences, and because each case contributes a matched model-clinician pair. We additionally computed Cohen's $d$ for each feature, a standardized effect size that expresses the mean paired difference in units of the within-case standard deviation, to distinguish statistically significant differences from practically large ones. To summarize overall stylistic proximity, we defined a standardized linguistic distance as the mean of $|$model $-$ clinician$|$ / SD$_{\mathrm{clinician}}$ across all features, where SD$_{\mathrm{clinician}}$ is the between-case standard deviation of the clinician values.

We applied this analysis to two configurations: the submitted two-step pipeline and the exploratory three-step rewrite variant. To visualize category-level profiles, we produced a radar chart with six composite scores, one per feature category, normalized so that the clinician profile traces a unit circle (Figure~\ref{fig:radar}). Each spoke value is the mean of per-feature model-to-clinician ratios within that category. Values above 1.0 indicate that the model exceeds the clinician on that dimension; values below 1.0 indicate that it falls short.

\subsection{Sensitivity Analyses}

\textbf{Task decomposition ablation.} The linguistic analysis revealed a stylistic gap between model outputs and clinician-authored answers, suggesting that generating factually grounded, stylistically appropriate, and length-constrained text within a single prompt creates competing objectives. To test whether separating these objectives narrows the gap, we compared the submitted pipeline against a three-stage \textit{rewrite} variant that (1) generates an unconstrained answer, (2) rewrites it for clinical register, and (3) enforces the 75-word limit as a separate call. The rewrite variant was implemented after the submission deadline. We evaluated word count compliance and standardized linguistic distance to the clinician reference as outcome measures.

\section{Results}
\label{sec:results}

\subsection{Subtask~2: Evidence Identification}

The answer-first ensemble achieved Strict Micro F1 62.90 on the test set. On the development set, the best single-run configuration achieved F1 66.22. For the test submission, we combined seven configurations by majority vote, predicting a sentence as Essential if at least four of seven runs classified it as such. Table~\ref{tab:subtask2} shows development and test results.

\begin{table*}[!ht]
\centering
\small
\begin{tabular}{lrrrrrr}
\hline
 & St.~P & St.~R & St.~F1 & Le.~P & Le.~R & Le.~F1 \\
\hline
Dev (best) & 55.62 & 81.82 & 66.22 & 62.66 & 81.82 & 70.97 \\
Test (4/7) & 59.30 & 66.96 & \textbf{62.90} & 74.27 & 66.96 & 70.43 \\
\hline
\end{tabular}
\caption{Subtask~2 results for the best configuration (development) and the majority-vote ensemble (test). St.\ = Strict, Le.\ = Lenient. All metrics are micro-averaged.}
\label{tab:subtask2}
\end{table*}

The ensemble outperformed the best individual test submission by 3.2 F1 points. Development recall (81.82\%) exceeded test recall (66.96\%) by 14.9 percentage points, suggesting distribution shift on the larger 47-case test set.

\subsection{Subtask~3: Answer Generation}

The answer-first pipeline achieved test Overall 31.90 (development Overall 32.21). Table~\ref{tab:subtask3} shows full results for the submitted configuration.

\begin{table*}[!ht]
\centering
\small
\begin{tabular}{lrrrrrrrr r}
\hline
 & AS & BS & BLEU & MC & R-1 & R-2 & R-L & SARI & Ov. \\
\hline
Dev  & 25.00 & 41.89 & 4.01 & 41.30 & 39.17 & 11.24 & 22.29 & 58.77 & 32.21 \\
Test & 24.53 & 41.80 & 6.12 & 37.52 & 38.84 & 11.95 & 24.18 & 57.24 & \textbf{31.90} \\
\hline
\end{tabular}
\caption{Subtask~3 results (Overall = mean of all reported metrics). AS = AlignScore, BS = BERTScore, MC = MEDCON, R = ROUGE, Ov.\ = Overall.}
\label{tab:subtask3}
\end{table*}

SARI was the highest-scoring metric on both sets (58.77 development, 57.24 test), reflecting simplification quality relative to the source note. BLEU was the lowest (4.01 development, 6.12 test), consistent with the observation that model outputs differed in surface form from the reference answers even when semantic similarity (BERTScore 41.89, AlignScore 25.00) and concept coverage (MEDCON 41.30) were moderate on the development set.

\subsection{Subtask~4: Answer-Evidence Alignment}

Self-consistency voting achieved test F1 79.81 on Subtask~4 (development F1 86.0). We drew $N = 5$ independent samples per case and retained links appearing in at least $t = 3$ of 5 samples. Table~\ref{tab:subtask4} shows development and test results.

\begin{table}[!ht]
\centering
\small
\begin{tabular}{lrrr}
\hline
 & P & R & F1 \\
\hline
Development & 85.1 & 87.0 & 86.0 \\
Test        & \textbf{83.6} & \textbf{76.3} & \textbf{79.8} \\
\hline
\end{tabular}
\caption{Subtask~4 results (P = precision, R = recall, F1 = micro F1).}
\label{tab:subtask4}
\end{table}

The 6.2-point development-to-test F1 drop (86.0 to 79.8) reflects the larger and more diverse test set (147 cases, including 127 cases from the 2025 dataset). Across all five teams on the official leaderboard, precision ranged from 82 to 88 and recall from 71 to 78, indicating that recall was the primary differentiator in this task.

We selected $N = 5$ and $t = 3$ based on practical and empirical considerations. Five samples balanced prediction diversity against API cost, representing a 5$\times$ increase over a single call. Lowering $t$ from 4 to 3 recovered 4.2 percentage points of recall at a cost of 2.4 points of precision, yielding a net F1 gain of 1.34 points on the development set.

\subsection{Error Analysis}

Iterative correction resolved all constraint violations across 40 experiments (1,171 total cases: 701 development, 470 test). The sections below characterize correction effectiveness and the distribution of failure modes.

\subsubsection{Correction Effectiveness}

The correction loop achieved 100\% compliance on both the development and test sets (see Appendix Table~\ref{tab:compliance} for per-attempt rates). First-try compliance was 21.6 percentage points lower on the test set (64.3\% versus 85.9\%), reflecting longer clinical notes and unseen question types. The correction loop closed this gap fully within five attempts in both conditions.

\subsubsection{Error Taxonomy}

\begin{figure*}[!ht]
\centering
\includegraphics[width=\textwidth]{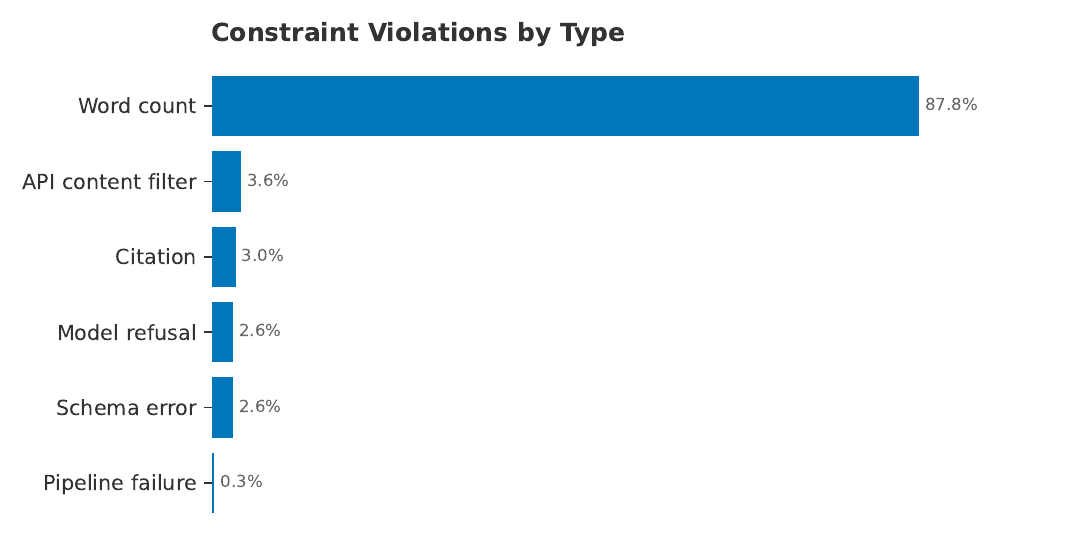}
\caption{Distribution of first-attempt error types on the development ($n = 129$ errors) and test ($n = 177$ errors) sets. Word count violations dominated both conditions, rising from 75.2\% of development errors to 94.9\% of test errors.}
\label{fig:errors}
\end{figure*}

Word count violations accounted for 75.2\% of development errors and 94.9\% of test errors. Citation errors, schema errors, and pipeline failures were absent from the test set entirely, indicating that development-set tuning eliminated structural and citation errors but did not generalize to length constraint adherence on longer test notes. Figure~\ref{fig:errors} shows the full breakdown.

\subsubsection{Content Safety}

Two distinct content safety failure modes affected a small fraction of cases. Infrastructure-level content filtering (non-deterministic): the API silently returned no response, causing a downstream error. Eleven unique cases were affected (4 development, 7 test); retries typically succeeded. Model refusal (persistent): the model returned safety-decline text that could not be parsed as a structured response. Eight unique cases were affected (7 development, 1 test), all involving sensitive clinical content. Because these events were input-dependent, they could not be resolved by correction prompts; multi-candidate pipelines ($k > 1$) provided partial resilience by allowing other candidates to succeed even when one call was refused.

\subsection{Linguistic Analysis}

\subsubsection{Model versus Clinician Divergence}

Model outputs diverged from clinician-authored reference answers on 8 of 45 linguistic features (Wilcoxon signed-rank, Benjamini-Hochberg adjusted $p < 0.05$). Table~\ref{tab:linguistic} (Appendix) lists all features for the submitted configuration, sorted by effect size (Cohen's $d$).

Three divergence patterns dominated. First, model outputs were harder to read across all three readability metrics. The Flesch Reading Ease score was 21 points lower (46.7 versus 25.6, $d = 1.11$), the Flesch-Kincaid grade was 3.2 points higher (10.7 versus 13.9, $d = -0.93$), and the Coleman-Liau grade was 4.1 points higher (12.8 versus 16.9, $d = -1.12$). Second, models used longer individual words (5.39 versus 6.12 characters, $d = -1.21$) and fewer function words (40.9\% versus 34.2\%, $d = 1.26$), producing denser prose with higher noun and adjective rates. Third, models produced more lexically diverse outputs than clinicians, with higher type-token ratios (0.756 versus 0.791, $d = -0.60$) and higher hapax ratios (0.599 versus 0.670, $d = -0.77$).

Notably, word count (73.5 versus 72.7, $p = 0.085$) and sentence count (4.75 versus 4.45, $p = 0.303$) did not differ significantly. The submitted configuration brought word count and sentence count within the range of clinician values, resolving the length and sentence structure divergences seen in earlier configurations.

\subsubsection{Two-Step versus Three-Step Linguistic Distance}

\begin{figure*}[!ht]
\centering
\includegraphics[width=\textwidth]{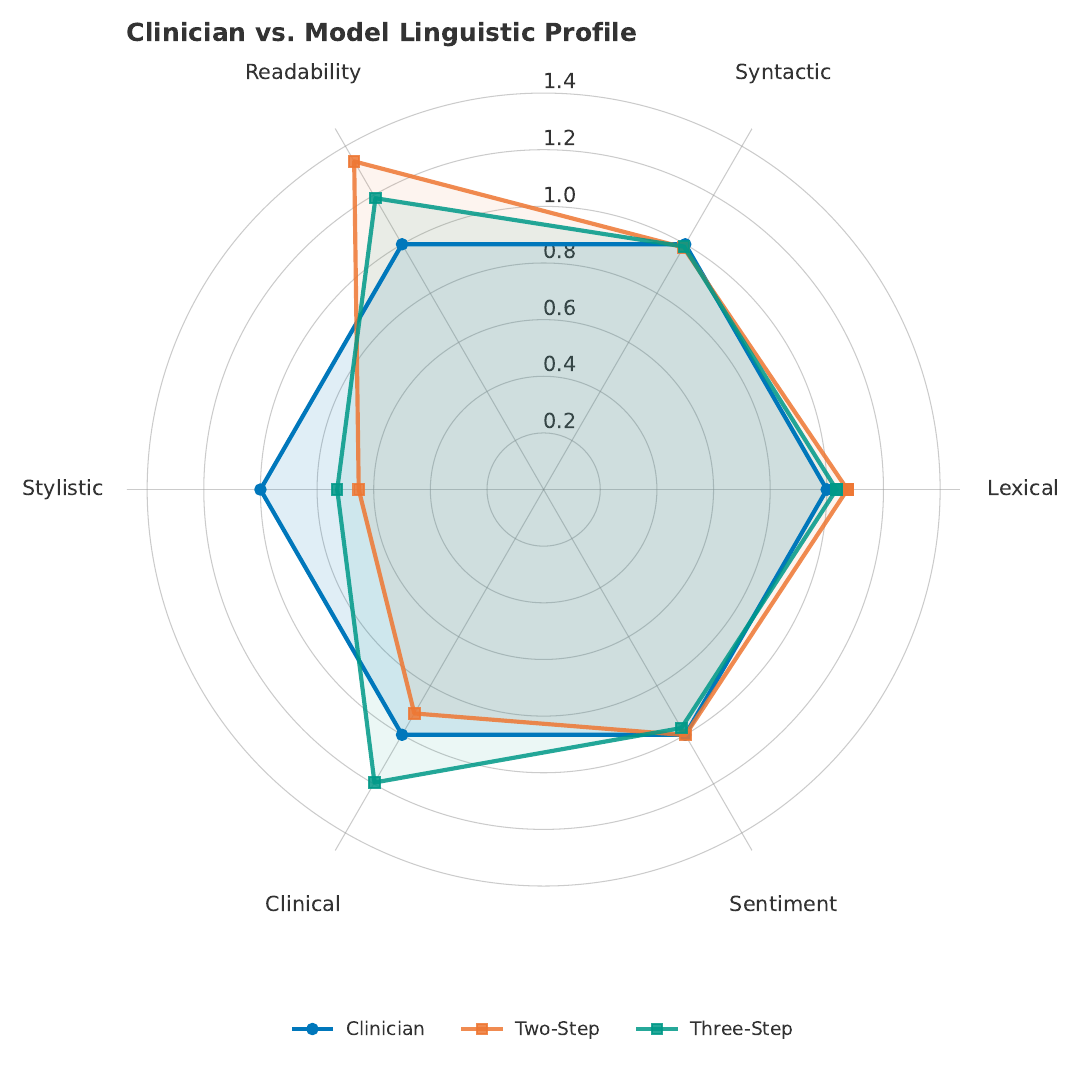}
\caption{Category-level linguistic profiles for the Two-Step and Three-Step configurations. The clinician reference traces the unit circle. Spoke values are the mean of per-feature model-to-clinician ratios within each category, with all features weighted equally; values above 1.0 indicate the model exceeds the clinician on that dimension, and values below 1.0 indicate the model falls short.}
\label{fig:radar}
\end{figure*}

The three-step rewrite pipeline reduced linguistic distance from 0.519 to 0.381 and improved Subtask~3 Overall by approximately 2 points on the development set. Table~\ref{tab:linguistic_dist} shows the full comparison; Figure~\ref{fig:radar} shows category-level profiles as a radar chart normalized to the clinician reference.

\begin{table}[!ht]
\centering
\footnotesize
\begin{tabular}{lrrr}
\hline
\textbf{Feature} & \textbf{Clinician} & \textbf{3-Step} & \textbf{2-Step} \\
\hline
Linguistic distance    & --    & \textbf{0.381} & 0.519 \\
Dev Overall (ST3)      & --    & \textbf{34.09} & 32.21 \\
Word count             & 73.5  & 73.4  & 72.7  \\
Sentence count         & 4.75  & 4.65  & 4.45  \\
Avg sentence length    & 17.98 & 19.17 & 19.88 \\
Flesch-Kincaid grade   & 10.72 & 12.52 & 13.87 \\
Passive voice ratio    & 0.39  & 0.29  & 0.24  \\
``the patient'' count  & 1.55  & 0.95  & 1.10  \\
Total hedging          & 0.50  & 0.45  & 0.40  \\
Temporal connectives   & 0.35  & 0.60  & 0.80  \\
\hline
\end{tabular}
\caption{Linguistic distance and selected features for the Two-Step (submitted) and Three-Step (post-deadline) configurations. ST3 = Subtask~3.}
\label{tab:linguistic_dist}
\end{table}

Improvements were distributed across all six feature categories. The most pronounced gains were in readability (Flesch-Kincaid grade 12.52 versus 13.87, closer to the clinician's 10.72) and syntactic structure (sentence count 4.65 versus 4.45; average sentence length 19.17 versus 19.88 words). Passive voice ratio remained below the clinician reference in both configurations, and ``the patient'' frequency was below clinician levels in both (0.95 for Three-Step, 1.10 for Two-Step, versus clinician 1.55). The Three-Step pipeline achieved closer stylistic alignment and also improved Subtask~3 automatic metrics: Dev Overall 34.09 versus 32.21 for the Two-Step configuration, a gain of approximately 2 points.

\section{Discussion}

\subsection{Main Findings}

LLMs equipped with answer-grounded evidence classification, iterative constraint enforcement, and self-consistency voting can achieve competitive performance on grounded EHR question answering. These results demonstrate that the answer-first paradigm improved both evidence selection accuracy and output compliance compared to direct classification baselines. Iterative correction resolved all remaining constraint violations in our experiments. At the same time, our linguistic analysis reveals that even configurations that perform competitively on automatic metrics produce text that is measurably harder to read and stylistically less similar to clinician-authored answers than word count and sentence count alone would suggest.

\subsection{Evidence Identification (Subtask~2)}

The ensemble submission ranked third on evidence identification (Strict Micro F1 62.90), gaining approximately 3 F1 points over the best single-run submission (59.69). These gains support the central hypothesis that evidence classification is easier after generating a candidate answer than before: the model's own prior generation provides an explicit context for judging which sentences were relied upon, reducing the difficulty of the minimal-set determination.

The test F1 of 62.90 places our system within a competitive range. Our best submission achieved higher recall than precision (P = 59.30, R = 66.96), consistent with the observation that under-retrieval is more costly than over-retrieval in this task.

The configuration sweep suggested that pipeline method had a larger effect on F1 than model version alone. However, reasoning effort and candidate count ($k$) varied across runs, preventing clean isolation of individual factor contributions. Ensemble diversity, spanning both model versions and multiple $k$ settings, was the primary driver of the test-set gain over single-run baselines.

\subsection{Answer Generation (Subtask~3)}

The Subtask~3 submission achieved an overall score of 31.90. Interpreting this result requires caution because the official overall score aggregates nine heterogeneous metrics (lexical overlap, semantic similarity, medical concept coverage, and simplification quality) that may partially conflict. SARI was the highest-scoring individual metric (57.24), reflecting simplification quality relative to the clinical note source.

The linguistic analysis in Section~\ref{sec:results} reveals an important gap between metric performance and stylistic register. Although the submitted configuration eliminated statistically significant word count and sentence count divergences from the clinician reference, 8 of 45 linguistic features remained significantly different after Benjamini-Hochberg correction. Model outputs were consistently harder to read (Flesch-Kincaid grade 3.2 points higher), used longer individual words (6.12 versus 5.39 characters), and contained fewer function words (34.2\% versus 40.9\%). These properties reflect a tendency toward technical, noun-heavy prose that may be less accessible to lay readers despite matching the reference on length. Whether this stylistic divergence matters for clinical utility cannot be determined from automatic metrics alone and requires human or clinical evaluation.

\subsection{Answer-Evidence Alignment (Subtask~4)}

Self-consistency voting with $N = 5$ samples per case and a fixed threshold of $t = 3$ achieved test F1 79.81 on Subtask~4. The vote-count decomposition on the development set showed a clear monotonic relationship between vote count and precision (approximately 80 to 90\% at $t = 5$, approximately 55 to 70\% at $t = 3$, approximately 15 to 20\% at $t = 1$), motivating the threshold choice. Lowering $t$ from 4 to 3 recovered 4.2 percentage points of recall (72.1\% to 76.3\%) while reducing precision by 2.4 points (86.0\% to 83.6\%), yielding a net F1 gain of 1.34 points.

The full leaderboard pattern for Subtask~4 indicates that all five participating teams were precision-heavy (P = 82 to 88, R = 71 to 78), and the spread in recall largely determined the ranking. This pattern suggests that recall is the primary differentiator in the answer-evidence alignment task. The development-to-test drop (F1 86.0 to F1 79.8) reflects the larger and more diverse test set: 127 of the 147 test cases came from the 2025 challenge dataset.

\section{Conclusion}

We described the UIC-AIHealth4All system for ArchEHR-QA 2026, which participated in three subtasks of grounded question answering from electronic health records. Our primary contributions are an answer-first evidence classification pipeline for Subtasks~2 and 3, an iterative constraint enforcement loop that achieved full output compliance in our experiments, and self-consistency voting for Subtask~4 alignment.

The answer-first pipeline enabled an ensemble that ranked third on evidence identification (Strict Micro F1 62.90). The iterative correction loop achieved 100\% constraint compliance on both the development and test sets within five attempts, demonstrating robustness to the distribution shift introduced by longer and more diverse test notes. Self-consistency voting with a fixed threshold of $t = 3$ ranked fifth on answer-evidence alignment (F1 79.81), with vote count providing a reliable proxy for link precision.

The linguistic analysis reveals a persistent stylistic gap between model outputs and clinician-authored answers that is not captured by the official evaluation metrics. Despite prompt engineering that successfully matched clinician word count and sentence count, model outputs remained substantially harder to read, used longer and less functional vocabulary, and showed fewer pronouns and passive constructions than the clinician reference. The post-deadline three-step rewrite pipeline narrowed this gap (linguistic distance 0.381 versus 0.519 for the submitted two-step pipeline) at a 9.5-fold latency cost. This result suggests that readability and stylistic register are separable from factual accuracy and that decomposed prompting can address them, but at a computational cost that may not be acceptable in production systems.

Future work should include human and clinical evaluation to determine whether the residual stylistic divergences affect patient understanding or clinical utility. Applying the answer-first paradigm to other grounded retrieval tasks outside the clinical domain would test the generality of the core hypothesis. Developing efficient alternatives to multi-stage decomposition that reduce the latency cost of register correction remains an open problem. Finally, extending the self-consistency voting approach to evidence classification and answer generation---not only alignment---may further improve recall without sacrificing the precision gains achievable through threshold tuning.

\section{Limitations}

We conducted no human or clinical evaluation. Automatic metrics may not capture dimensions of answer quality most relevant to patients or clinicians, such as appropriateness, emotional tone, or clinical accuracy.

The multi-stage pipeline requires at least three sequential API calls per case. The three-step rewrite variant increases latency by a factor of 9.5, which may preclude deployment in time-sensitive clinical settings.

All experiments used a single dataset (ArchEHR-QA 2026) derived from the MIMIC-III and MIMIC-IV databases. Performance may not generalize to other EHR systems, note types, clinical specialties, or patient populations.

The system depends on proprietary models (GPT-5.1 and GPT-5.2 through Azure OpenAI). Results may not transfer to open-weight models, and API behavior may change across versions without notice.

The linguistic analysis was conducted on 20 development cases, limiting statistical power for features with small effect sizes. Results for the strongest divergences (readability metrics, $d > 0.9$) are robust, but findings for features near the significance threshold should be interpreted with caution.

\section{Ethical Considerations}

All clinical notes were accessed through credentialed PhysioNet accounts under a signed data use agreement.\footnote{\url{https://physionet.org/news/post/gpt-responsible-use}} We developed the system using Azure OpenAI Services under a university account with a Business Associate Agreement (BAA). We applied for and received modified abuse monitoring, which disables human review of prompts and completions by the API provider. No protected health information was used during prompt development or tuning.

\section{Acknowledgements}

This work was supported by the AI.Health4All Center at the University of Illinois Chicago.  Natalie Parde was also supported by the National Institutes of Health under Grants No.~R41NR020667, 1R61DA057629-01A1, and 1R01AG091762-01 and the National Science Foundation under Grant No.~2125411 during this time.  Any opinions, findings, and conclusions or recommendations are those of the authors and do not necessarily reflect the views of the National Science Foundation or the National Institutes of Health.

\section{Bibliographical References}\label{sec:reference}
\bibliographystyle{lrec2026-natbib}
\bibliography{references}

\label{lr:ref}

\appendix

\section{Correction Effectiveness}

\begin{table}[!ht]
\centering
\small
\begin{tabular}{lrr}
\hline
 & \textbf{Dev} ($n = 701$) & \textbf{Test} ($n = 470$) \\
\hline
After attempt 1 & 85.9\% & 64.3\% \\
After attempt 2 & 90.9\% & 72.8\% \\
After attempt 3 & 96.1\% & 86.2\% \\
After attempt 4 & 98.1\% & 91.1\% \\
After attempt 5 & 99.4\% & 98.1\% \\
Final            & 100.0\% & 100.0\% \\
\hline
\end{tabular}
\caption{Cumulative constraint compliance by correction attempt across 35 development and 10 test experiments.}
\label{tab:compliance}
\end{table}

\section{Linguistic Feature Analysis}

\begin{table*}[!ht]
\centering
\small
\begin{tabular}{lrrrrr}
\hline
\textbf{Feature} & \textbf{Clinician} & \textbf{Model} & \textbf{d} & \textbf{$p_{\mathrm{adj}}$} & \\
\hline
Function word \%           & 40.9\% & 34.2\% & 1.26  & $<$0.001 & *** \\
Average word length        & 5.39   & 6.12   & -1.21 & $<$0.001 & *** \\
Coleman-Liau grade         & 12.8   & 16.9   & -1.12 & $<$0.001 & *** \\
Flesch Reading Ease        & 46.7   & 25.6   & 1.11  & $<$0.001 & *** \\
Flesch-Kincaid grade       & 10.7   & 13.9   & -0.93 & 0.005    & **  \\
Hapax ratio                & 0.599  & 0.670  & -0.77 & 0.022    & *   \\
Adjective rate             & 8.2\%  & 10.7\% & -0.75 & 0.018    & *   \\
Hapax count                & 44.0   & 48.7   & -0.72 & 0.031    & *   \\
Pronoun rate               & 6.3\%  & 3.4\%  & 0.69  & 0.080    & ns  \\
Type-token ratio           & 0.756  & 0.791  & -0.60 & 0.085    & ns  \\
Temporal connectives       & 0.35   & 0.80   & -0.59 & 0.085    & ns  \\
Noun rate                  & 31.0\% & 34.6\% & -0.56 & 0.080    & ns  \\
Passive voice ratio        & 0.39   & 0.24   & 0.55  & 0.085    & ns  \\
MTLD                       & 87.0   & 100.8  & -0.54 & 0.085    & ns  \\
Third-person pronouns      & 3.45   & 1.85   & 0.53  & 0.085    & ns  \\
Causal connectives         & 0.90   & 0.35   & 0.52  & 0.100    & ns  \\
Medical abbreviations      & 0.95   & 1.65   & -0.51 & 0.087    & ns  \\
Unique word count          & 55.5   & 57.5   & -0.49 & 0.103    & ns  \\
First-person pronouns      & 0.15   & 0.00   & 0.41  & n/a      & n/a \\
Avg sentence length (words)& 17.98  & 19.88  & -0.38 & 0.142    & ns  \\
Sentence length SD         & 6.52   & 5.14   & 0.37  & 0.255    & ns  \\
Word count                 & 73.5   & 72.7   & 0.36  & 0.159    & ns  \\
Budget utilization         & 0.980  & 0.969  & 0.36  & 0.142    & ns  \\
Sentence opener diversity  & 0.973  & 1.00   & -0.31 & n/a      & n/a \\
``the patient'' count      & 1.55   & 1.10   & 0.31  & 0.344    & ns  \\
Sentence count             & 4.75   & 4.45   & 0.26  & 0.496    & ns  \\
Adposition ratio           & 10.9\% & 10.1\% & 0.25  & 0.576    & ns  \\
Determiner ratio           & 8.5\%  & 7.7\%  & 0.23  & 0.618    & ns  \\
Certainty markers          & 0.60   & 0.45   & 0.20  & 0.726    & ns  \\
Adverb rate                & 2.4\%  & 2.9\%  & -0.19 & 0.576    & ns  \\
Subjectivity               & 0.380  & 0.426  & -0.17 & 0.952    & ns  \\
Epistemic hedging          & 0.45   & 0.35   & 0.13  & 0.952    & ns  \\
Sentiment polarity         & 0.028  & 0.017  & 0.12  & 0.576    & ns  \\
Warning expressions        & 0.85   & 1.10   & -0.12 & 0.969    & ns  \\
Total hedging              & 0.50   & 0.40   & 0.11  & 0.952    & ns  \\
Contrastive connectives    & 0.20   & 0.25   & -0.10 & 1.0      & ns  \\
Verb rate                  & 13.7\% & 13.5\% & 0.04  & 0.969    & ns  \\
Total connectives          & 1.45   & 1.40   & 0.03  & 0.867    & ns  \\
Average parse depth        & 5.30   & 5.32   & -0.01 & 1.0      & ns  \\
Hedge-certainty ratio      & 0.458  & 0.458  & 0.00  & 0.969    & ns  \\
Reassurance expressions    & 0.60   & 0.60   & 0.00  & 1.0      & ns  \\
Evidential hedging         & 0.05   & 0.05   & 0.00  & n/a      & n/a \\
Meta-referential exprs     & 0.00   & 0.00   & 0.00  & n/a      & n/a \\
Non-ASCII characters       & 0.00   & 0.00   & 0.00  & n/a      & n/a \\
Second-person pronouns     & 0.00   & 0.00   & 0.00  & n/a      & n/a \\
\hline
\end{tabular}
\caption{All 45 linguistic features for the submitted configuration versus the clinician reference, sorted by Cohen's $d$ magnitude. $p_{\mathrm{adj}}$: Benjamini-Hochberg adjusted $p$-values (Wilcoxon signed-rank test). Significance: *** $p_{\mathrm{adj}} < 0.001$, ** $p_{\mathrm{adj}} < 0.01$, * $p_{\mathrm{adj}} < 0.05$, ns = not significant, n/a = test not applicable.}
\label{tab:linguistic}
\end{table*}

\end{document}